\documentclass[a4,11pt]{article}

\usepackage[style=numeric]{biblatex}

\usepackage{adjustbox,import,float,epstopdf}
\usepackage{amsfonts}
\usepackage{amsmath}
\usepackage[small,compact]{titlesec}
\newcommand{\squishlist}{
   \begin{list}{$\bullet$}
    { \setlength{\itemsep}{0pt}    \setlength{\parsep}{0pt}
      \setlength{\topsep}{0pt}     \setlength{\partopsep}{0pt}
      \setlength{\leftmargin}{2em} \setlength{\labelwidth}{1.5em}
      \setlength{\labelsep}{0.5em} } }

\newcommand{\squishend}{
    \end{list}  }

\usepackage{fancyhdr}
\usepackage{times}
\usepackage{url}

\makeatother

\usepackage{wrapfig}
\usepackage{lipsum}
\usepackage{tikz}
\usepackage{hyperref}
\usepackage{tabularx}
\usepackage{booktabs}
\usepackage{multicol}

\usepackage{caption}
\begin{document}

\setcounter{page}{1}

\begin{center}
    \LARGE \textbf{Federated learning for unpaired multimodal data } \\[0.5em]

        \normalsize Anders Eklund \\ Department of Biomedical Engineering, Linköping University, Sweden \\ Department of Computer and Information Science, Linköping University, Sweden 
    \\ Center for Medical Image Science and Visualization (CMIV), Linköping University, Sweden

\end{center}

\vspace{1.5cm}

\vspace{-50pt}
\section*{Introduction}
 
\begin{wrapfigure}[15]{r}{0.6\textwidth}
  \begin{center}
    \vspace{-30pt}
    \includegraphics[width=0.6\textwidth]{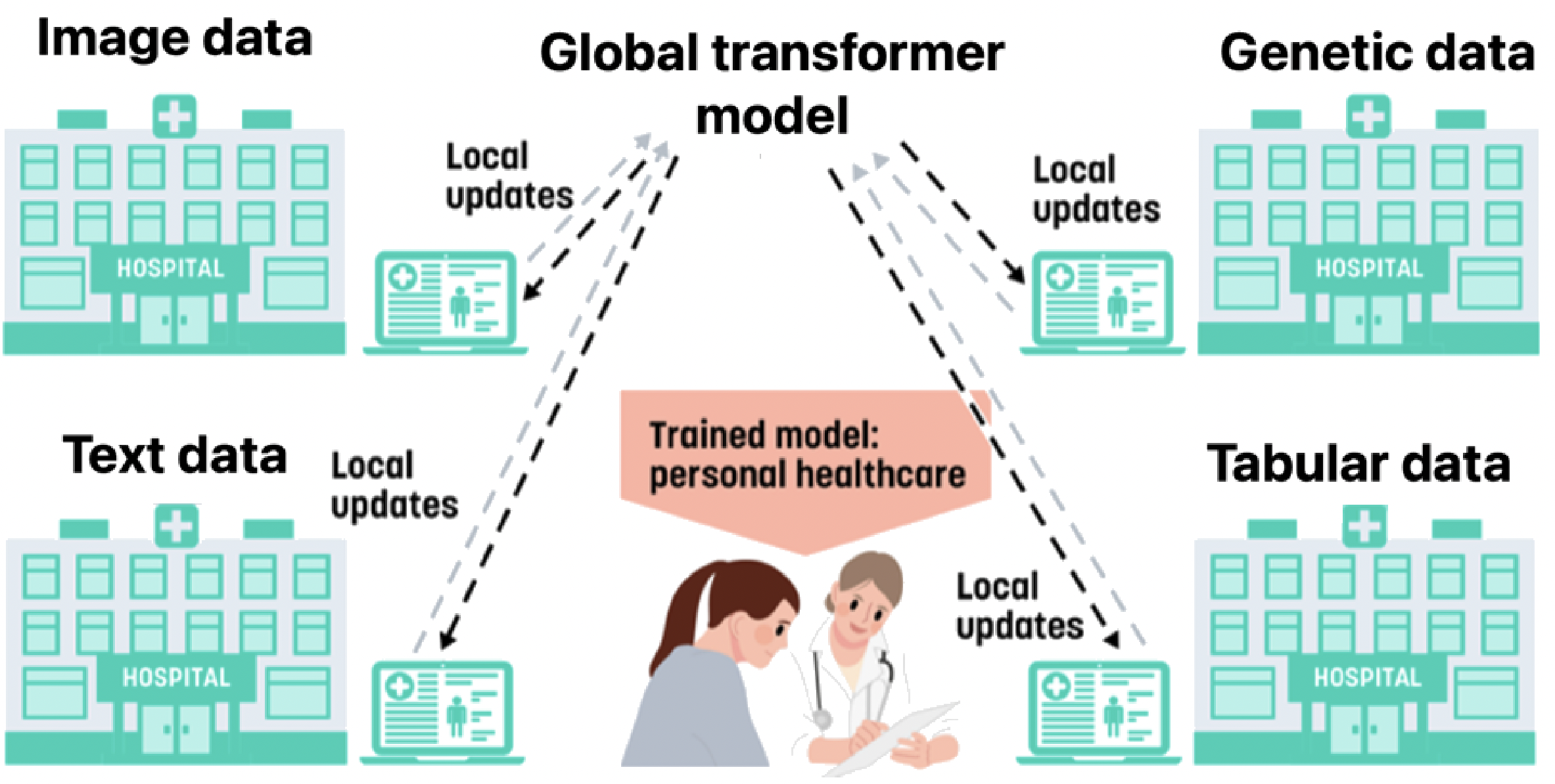}
  \end{center}
  \vspace{-15pt}
  \caption{We propose novel federated methods for \emph{unpaired} multimodal data, where each node only has access to one modality (e.g. images, text, genetics, tabular data). As all data types can be tokenized, they can be fed into the same transformer model.}
  \label{radiotherapy}
\end{wrapfigure}

There are many domains where the data are sensitive (e.g., medical applications, financial applications, software, patents, employee/customer data, etc.)\ and where privacy laws and policies make it difficult to centrally store large datasets to train large machine learning models on. \textbf{Federated learning} (FL) offers the only viable solution to both data privacy and training with large datasets, by keeping the data locally at each participating site (node). In FL a global model is trained by repeatedly combining models from a number of nodes. The model at each node is only trained using the local data at that node (Figure~\ref{radiotherapy}). Recent models (e.g. vision language models (VLM)) rely on multimodal data to improve performance compared to unimodal data~\cite{acosta2022multimodal,zhang2024vision, gupta2025better}. \textbf{However, multimodal models are normally trained using \emph{paired} multimodal data, and such datasets are often hard to create.} This is especially true in the medical

 domain, where \emph{paired} multimodal data means that all modalities must be available for all patients. Paired multimodal medical data are also problematic from an anonymization perspective, as more modalities (e.g. images, text, genetics, tabular data) increase the probability of identifying a patient. \textbf{Here, we will therefore propose federated methods for \emph{unpaired} multimodal data, to handle the scenario where each node in a federation only has one type of data. This is achieved by feeding tokens from different \emph{unpaired} modalities into the same global transformer model~\cite{gupta2025better, edamadaka2025universally} (Figure~\ref{radiotherapy}).}

\section*{State-of-the art} 

According to Tables~\ref{tab:fl_survey} and~\ref{tab:fl_survey2} it is clear that our proposed approach is the first to combine a homogenous multimodal transformer model, low-rank approximations (LoRA and DoRA), modality alignment regularization through centered kernel alignment (CKA), uncertainty-weighted and layer-wise aggregation, for communication efficient FL with unpaired multimodal data (while also improving privacy). \textbf{A key distinction of our approach is the unification of disjoint modalities into a shared homogenous transformer (motivated by the recent platonic representation hypothesis~\cite{maniparambil2024vision,huh2024platonic,edamadaka2025universally}).} This hypothesis states that any model trained for a certain task will converge to similar embeddings and representations of the world, regardless of the type of data (modality) used to train the model. Previous methods for multimodal FL typically employ modality-specific architectures (e.g., convolutional neural networks (CNNs) for images, recturrent neural networks (RNNs) for text), which precludes parameter-efficient aggregation strategies like LoRA due to the lack of a single homogenous model. Compared to~\cite{gupta2025better,edamadaka2025universally} we propose methods for multimodal unpaired \emph{federated} learning, which is substantially harder compared to centralized learning with all data available in one node.

\begin{table}[htb]
\centering
\caption{Surveys about multimodal FL, regarding if they mention certain concepts or not.}
\label{tab:fl_survey}
\resizebox{\textwidth}{!}{%
\begin{tabular}{|l|l|l|l|l|l|}
\hline
\textbf{Paper / Feature} & Che 2023~\cite{che2023multimodal}  & Huang 2024~\cite{huang2024multimodal} & Trasher 2025~\cite{thrasher2025multimodal}  & Adam 2025~\cite{adam2025survey} &  This proposal \\ \hline
\textbf{Paper type} & Survey & Survey  & Survey & Survey &   Application\\ \hline

\textbf{Transformers} & Yes & Yes  & Briefly & Yes &  Yes \\ \hline 
\textbf{CKA} & No & No & No & No  &Yes\\ \hline
\textbf{LoRA} & No & No &  No&  Future direction&  Yes (\textbf{DoRA})\\ \hline

\textbf{Uncertainty aggregation} & No & No & No  & Briefly &   Yes \\ \hline 

\textbf{Layer-wise aggregation} & No & No & No  & No &   Yes \\ \hline 

\textbf{Adaptive LoRA rank} & No & No & No  & No &   Yes \\ \hline 

\textbf{Public dataset} & No & No & Yes & Yes &  Yes \\ \hline

\textbf{Starting model} & Yes & No (random) & No (random) & Future direction & VLM \\ \hline

\hline

\end{tabular}%
}
\end{table}



\begin{table}[htb]
\centering
\caption{Methods for (multimodal) FL, regarding if they use certain concepts or not. p. = paired}
\label{tab:fl_survey2}
\resizebox{\textwidth}{!}{%
\begin{tabular}{|l|l|l|l|l|l|l|}
\hline
\textbf{Paper / Feature} & Yu 2023~\cite{yu2023multimodal}&  Sun  2024~\cite{sun2024towards} & Wang 2024~\cite{wang2024flora} & Kim 2025~\cite{kim2025x} & This proposal \\ \hline

\textbf{Paper type} & FL multimodal & FL multimodal  & \textbf{FL for text only} & FL MM \textbf{locally p.} & Application \\ \hline

\textbf{Novelty}   & Contrastive reg.  & Multi-modal transf. & Fine tuning LoRA & Image synthesis & Several\\ \hline

\textbf{Transformers}   & No   & Yes & Yes & Yes& Yes multimodal \\ \hline 

\textbf{CKA}  & No  & No & No& No&Yes\\ \hline

\textbf{LoRA} & No   & No & Yes& Yes, \textbf{incorrectly} & Yes (\textbf{DoRA})\\ \hline

\textbf{Uncertainty aggregation}  &  No  & No & No & No& Yes \\ \hline 

\textbf{Layer-wise aggregation} & No & No & No  & No &   Yes \\ \hline 

\textbf{Adaptive LoRA rank} & No & No & Static  & No &   Yes, \textbf{dynamic} \\ \hline

\textbf{Alignment signal}  &  Embeddings (leaky)  & Prototypes (leaky) &  N/A (unimodal) & N/A (locally p. data)  & Gram m. (\textbf{private}) \\ \hline

\textbf{Public dataset}  & Yes   & No & No & No (locally p. data)& Yes \\ \hline

\textbf{Starting model}  & Random    & Random & LLM & VLM & VLM \\ \hline

\hline

\end{tabular}%
}
\end{table}

\section*{Methods}

\begin{figure}
  \begin{center}
    \vspace{-20pt}        \includegraphics[width=0.8\textwidth]{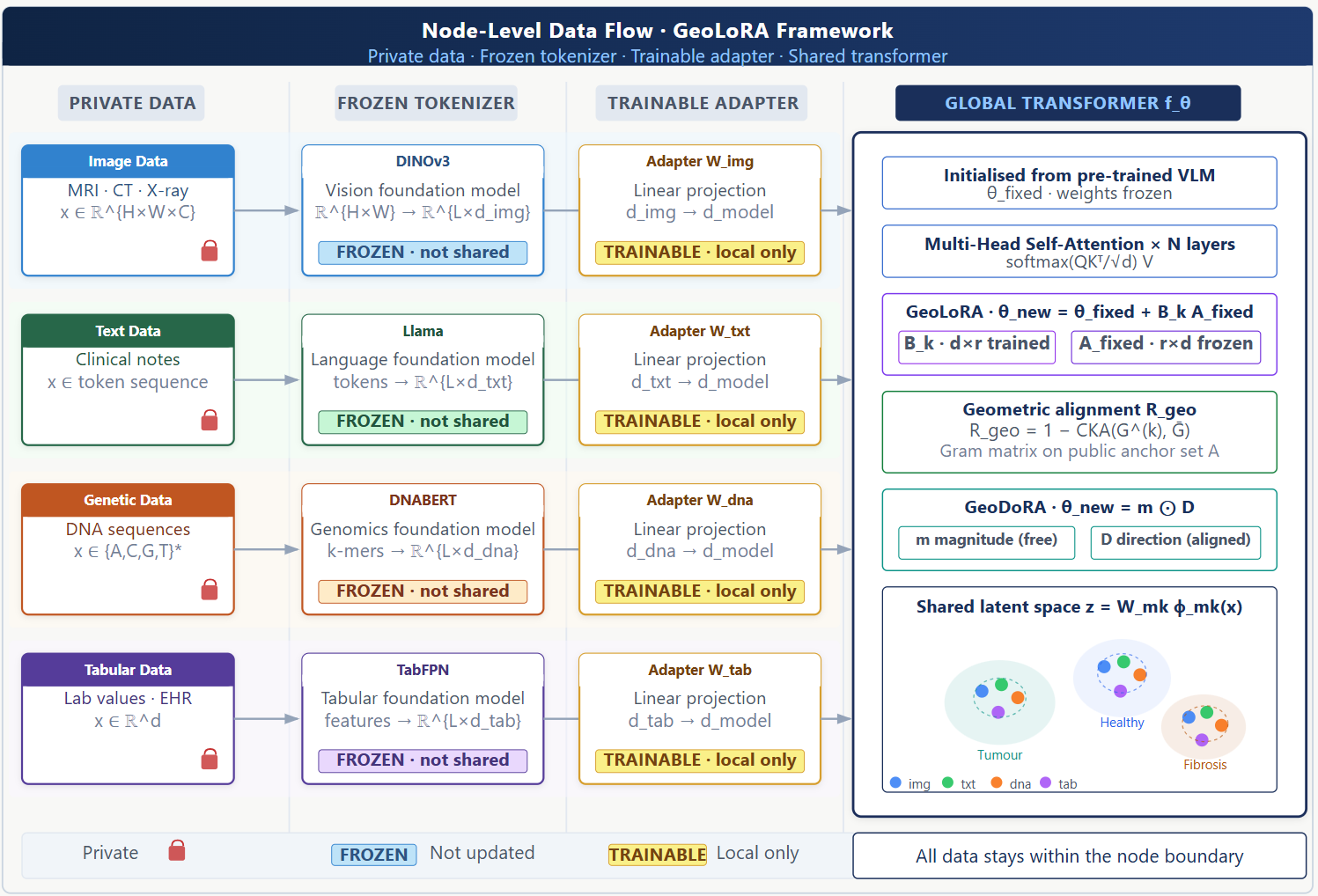}
  \end{center}
  \vspace{-15pt}
  \caption{Overview of the proposed approach. Data from unpaired modalities are converted to tokens (using pre-trained tokenizers) and then transformed to the same global dimension using modality specific adapters, to feed all modalities to a single global transformer model. Efficient updates of the huge transformer model are performed using low-rank approximations (LoRA). Centered kernel alignment is concurrently used with LoRA to geometrically align the different modalities through a public anchor dataset. Uncertainty metrics for each node are efficiently obtained by calculating distances to the anchor dataset in the latent space of the transformer.}
  \label{fig:overview}
\end{figure}

See Figure~\ref{fig:overview} for an overview of the proposed approach. To accelerate convergence and ensure robust feature extraction, each node in the federation will use a pre-trained tokenizer to project its local data into transformer tokens. This leverages the platonic convergence of foundation models, as recently demonstrated or proposed by several authors~\cite{maniparambil2024vision,huh2024platonic,edamadaka2025universally}, and will ensure that the input geometry is already partially aligned. Specifically, we will use DINOv3~\cite{simeoni2025dinov3} for image data, DNABERT~\cite{ji2021dnabert} for genetic data, TabFPN~\cite{hollmann2025accurate} for tabular data, and Llama~\cite{touvron2023llama} for text data. The tokenizers are frozen and not shared in the federation. To transform all modalities into the same dimension, each node will train an adapter which performs a (linear) projection from the tokenizer to the dimension of the global transformer model. The global transformer model will be initialized with pre-trained weights from a VLM.

We formalize our approach as a federated optimization problem over a set of $K$ nodes, $\mathcal{C} = \{1, \dots, K\}$, where each node $k$ possesses a private, unimodal dataset $\mathcal{D}_k = \{(x_i^{(k)}, y_i^{(k)})\}_{i=1}^{N_k}$, where $x_i$ is a data sample and $y_i$ is a label. The modality of node $k$ is denoted $m_k \in \{\mathcal{M}_{image}, \mathcal{M}_{text},  \mathcal{M}_{genetics}, \mathcal{M}_{tabular} \}$. Let $f_\theta(\cdot)$ be the shared global transformer model parameterized by $\theta$, operating on a latent dimension $d_{model}$ (same for all nodes). For node $k$ with modality $m_k$, we define a two-stage input processing pipeline:
\vspace{-0.2cm}
\begin{enumerate}
    \item \textbf{Frozen Tokenizer:} Let $\phi_{m_k}(\cdot)$ be a fixed, pre-trained tokenizer (e.g., DINOv3 or DNABERT) that maps the data $x$ to raw embeddings in $\mathbb{R}^{L \times d_{m_k}}$ ($L$ being the number of tokens). Note that $d_{m_k}$ varies by modality (node).
    \item \textbf{Trainable Adapter:} Let $W_{m_k} \in \mathbb{R}^{   d_{m_k} \times d_{model}} $ be a learnable linear projection matrix (``Adapter'') that maps the raw embeddings to the global transformer model dimension.
\end{enumerate}
\vspace{-0.2cm}
The unified token representation is thus $z = W_{m_k} \phi_{m_k}(x)$. Each node optimizes both the global transformer model $\theta$ and the local adapter $W_{m_k}$. Crucially, while $\theta$ is averaged across all nodes, $W_{m_k}$ is learned locally.

\begin{figure}
  \begin{center}
    \vspace{-20pt}        \includegraphics[width=0.9\textwidth]{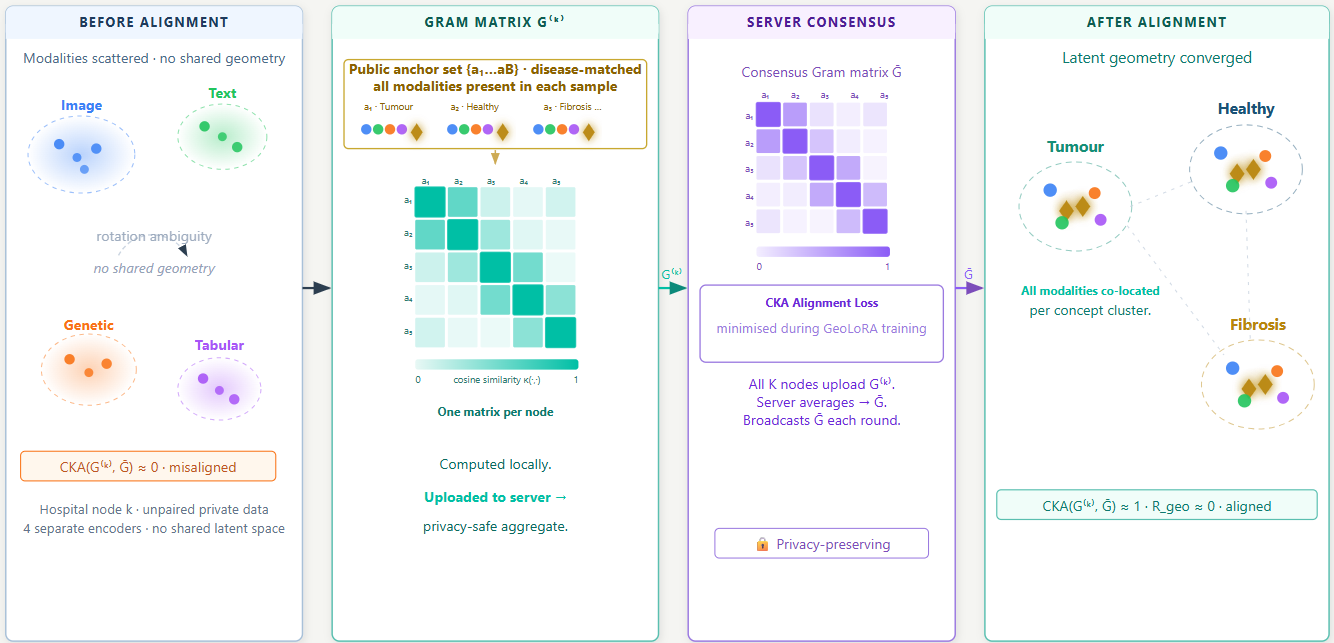}
  \end{center}
  \vspace{-15pt}
  \caption{Overview of the modality alignment approach. Without modality alignment the embeddings from different modalities for the same disease are far away from each other in latent space of the transformer. CKA is used on Gram matrices computed on the public anchor dataset, to quantify alignment between modalities. After modality alignment the embeddings from different modalities form clusters for different diseases. Only the Gram matrices computed on the public anchor dataset is transmitted between each node and the global server.}
  \label{fig:cka}
\end{figure}

\textbf{Idea 1: Alignment of unpaired modalities}. To enforce \textbf{representational convergence} between disjoint modalities, we introduce a public anchor set $\mathcal{A} = \{a_1, \dots, a_B\}$ of $B$ samples. Crucially, the modalities in this anchor dataset need not originate from the same patients (i.e. the dataset can be unpaired). The modalities must simply correspond to the same medical concept (i.e., originate from patients with the same disease). This allows us to align the latent manifolds of disjoint hospitals using only public, non-sensitive class prototypes, avoiding the privacy constraints of real patient pairing. One might ask: if a multimodal public anchor set exists, why not simply train the model centrally on that data? The answer lies in scale and domain specificity. Public medical datasets represent the tip of the iceberg, as they are often limited in size (e.g. $ N<10^3$) and biased toward textbook cases. In contrast, private hospital repositories contain the long tail of diverse pathologies and rare edge cases (e.g $N>10^6$) critical for clinical robustness. \textbf{We utilize the small public anchor set solely as a geometric Rosetta stone to align the manifolds}. In scenarios where a specific modality is missing from the public anchor set, we generate synthetic anchors (e.g., via digital twins or generative AI models). Crucially, our proposed precision-weighted aggregation (described later) naturally detects the distributional shift between real private data and synthetic anchors, assigning higher uncertainty to these nodes. During local training, node $k$ computes the Gram matrix $G^{(k)} \in \mathbb{R}^{B \times B}$ for a pair of samples $a_i, a_j$ in the public $\mathcal{A}$ dataset according to
\begin{equation}
    G_{ij}^{(k)} = \kappa \left( \text{Pool} \left( f_\theta \left( W_{m_k} \phi_{m_k}(a_i) \right) \right), \quad \text{Pool} \left(f_\theta \left( W_{m_k} \phi_{m_k}(a_j) \right) \right) \right),
\end{equation}
where $\kappa(\cdot, \cdot)$ is the cosine similarity kernel and $\text{Pool()}$ performs a pooling over $L$ tokens. See Figure~\ref{fig:cka} for a visualization of the modality alignment approach.

\textbf{We utilize the Gram matrix for three strategic reasons} 1) Sufficiency: As foundation models (e.g. our pre-trained tokenizers and global VLM) naturally converge toward shared representations~\cite{maniparambil2024vision,huh2024platonic,edamadaka2025universally}, aligning the relational geometry (Gram matrix) is sufficient to synchronize their coordinate systems without needing to retrain their features. 2) Efficiency: It offers a compact $O(B^2)$ representation that is significantly smaller than raw high-dimensional activations, minimizing communication overhead. 3) Privacy: While the anchor samples are public, sharing raw activations exposes the local model's specific feature space to model extraction attacks. The Gram matrix encapsulates only the relational geometry (scalar similarities), which prevents the server from reverse-engineering the private model parameters. As shown by Kornblith et al.~\cite{kornblith2019similarity}, the Gram matrix effectively captures the representational topology of deep networks. Our differentiable \textbf{geometric modality alignment loss} quantifies the spectral distance between the local kernel $G^{(k)}$ and the global consensus kernel $\bar{G}$ according to
\begin{equation}
    \mathcal{R}_{geo}(\theta) = \sum_{k=1}^K \left( 1 - \text{CKA}(G^{(k)}, \bar{G}) \right), \quad \text{CKA}(X, Y) = \frac{\text{tr}(XY^T)}{||X||_F ||Y||_F}.
\end{equation}

The global consensus kernel $\bar{G}$ serves as the target geometry for all nodes. It is computed by the central server by averaging the local kernel matrices $G^{(k)}$ uploaded by all nodes (not a privacy problem as the anchor dataset is public). In each federated round, the server broadcasts $\bar{G}$ to all nodes. This allows each node to align its local modality-specific manifold not just to another modality, but to the \emph{average biological geometry} of the entire federation. The penalty $\mathcal{R}_{geo}(\theta)$ forces the \textit{geometry} of the latent space to be invariant across modalities, ensuring that for example $f_\theta(\text{image})$ and $f_\theta(\text{text})$ map to the same manifold coordinates for semantically similar inputs (e.g. data from patients with the same disease). This is achieved through CKA~\cite{kornblith2019similarity}, \textbf{but unlike prior work limiting CKA to post-hoc analysis~\cite{kornblith2019similarity,maniparambil2024vision,edamadaka2025universally}, we uniquely repurpose it as an active regularization objective to steer the optimization.}  Standard CKA assumes a uniform, unsupervised alignment of the global manifold, which we empirically found leads to severe representation collapse when learning from highly imbalanced clinical data (see preliminary results). To protect the clinical structure of the minority class, we propose \textbf{supervised exponential CKA}. Let $M \in \mathbb{R}^{B \times B}$ be a binary supervision mask derived from the public anchor set, where $M_{ij} = 1$ if samples $a_i$ and $a_j$ share the same disease label, and $0$ otherwise. We apply an exponential temperature scaling $\tau_{align}$ to amplify local semantic neighborhoods while strictly penalizing inter-class overlap. Crucially, the exponential transformation is used here to create stronger gradients during backpropagation, ensuring that the critical but weak learning signals from the minority class are not washed out by the majority class. The modified, class-aware Gram matrix for node $k$ is formulated as

\begin{equation}
    \tilde{G}^{(k)} = \exp \left( \frac{G^{(k)}}{\tau_{align}} \right) \odot M,
\end{equation}

where $\odot$ denotes element-wise multiplication. The federated aggregation, to obtain the global model, is defined as a weighted sum of local losses $\sum_{k} p_k \mathcal{L}_k(\theta)$, regularized by the geometric alignment penalty as $\lambda \mathcal{R}_{geo}(\theta)$. 

While the primary innovation of our framework addresses strictly unpaired nodes, real-world clinical federations are often heterogeneous. Certain hospitals may possess locally paired multimodal datasets (e.g., patients with both a chest MRI and a corresponding textual radiology report). \textbf{Our framework natively capitalizes on these "hybrid nodes" by allowing them to act as structural bridge clients that rigidify the global manifold alignment.} For a hybrid node $h$ possessing a paired dataset $\mathcal{D}_{h} = \{(x_{m_1}^{(i)}, x_{m_2}^{(i)})\}_{i=1}^{N_h}$ for modalities $m_1$ and $m_2$, we introduce an auxiliary intra-node contrastive loss. Let $z_{m_1}^{(i)}$ and $z_{m_2}^{(i)} $ be the tokenized and projected latent representations for one paired multimodal sample $i$. We apply a symmetric InfoNCE loss~\cite{oord2018representation} to enforce direct geometric coupling between the paired modalities according to

\begin{equation}
\mathcal{L}_{bridge} = -\frac{1}{N_{B}} \sum_{i=1}^{N_{B}} \log \frac{\exp \left( \kappa \left( \text{Pool} \left( z_{m_1}^{(i)} \right), \text{Pool} \left(z_{m_2}^{(i)} \right) \right) /\tau_{c} \right)}   {\sum_{j=1}^{N_{B}} \exp \left( \kappa \left( \text{Pool} \left( z_{m_1}^{(i)} \right), \text{Pool} \left( z_{m_2}^{(j)} \right) \right)/\tau_{c} \right) },
\label{eq:bridge_loss}
\end{equation}

where $N_{B}$ is the batch size, $\kappa(\cdot, \cdot)$ is the cosine similarity, and $\tau_{c}$ is a contrastive temperature scalar. For these bridge clients, $\mathcal{L}_{bridge}$ is added to their local optimization objective alongside the task loss $\mathcal{L}_{task}$ and the public anchor geometric regularization $\mathcal{R}_{geo}$. By optimizing this joint objective, bridge clients upload low-rank updates (via GeoDoRA, described later) that contain exceptionally strong, explicitly paired semantic correlations. When aggregated at the central server, these updates act as rigid multi-modal anchor points within the global consensus kernel $\overline{G}$, drastically reducing the topological ambiguity for the strictly unpaired nodes.

\textbf{Idea 2: Efficiently updating a huge transformer model}. Leveraging the low intrinsic dimension of foundation model adaptation~\cite{huh2024platonic,edamadaka2025universally}, we exploit low-rank approximations (LoRA)~\cite{hu2022lora} to efficiently steer the massive pre-trained parameter space rather than retraining it, to reduce the required communication cost by \textbf{orders of magnitude (typically $>$ 99.9\%)}  (\cite{hu2022lora}) (shrinking the update for a foundation model from gigabytes to megabytes). To enable concurrent communication efficiency and geometric modality alignment, we propose \textbf{GeoLoRA}. GeoLoRA mathematically couples the low-rank updates to the shared manifold structure. We introduce low-rank matrices $A_k, B_k$ at node $k$ such that the local transformer model is updated as $ \theta_{new} = \theta_{fixed} + B_k A_k$.
We reformulate the local optimization objective to explicitly constrain the LoRA subspace using the geometric modality loss as

\begin{equation}
    \min_{W_{m_k}, A_k, B_k} \underbrace{\mathcal{L}_{task}(\mathcal{D}_k; W_{m_k}; \theta_{fixed} + B_k A_k)}_{\text{Task Performance}} + \lambda \underbrace{(1 - \text{CKA}(G^{(k)}_{adapted}, \bar{G}))}_{\text{Modality Alignment Regularization}},
\end{equation}
where $G^{(k)}_{adapted}$ is computed by passing the anchor set $\mathcal{A}$ through both the local adapter $W_{m_k}$ and the updated transformer model $\theta_{fixed} + B_k A_k $, ensuring the \textit{entire pipeline} aligns geometrically with the global consensus.


To ensure consistent aggregation of the low-rank updates across nodes, and to further reduce communication or enable a higher rank in LoRA, we employ a subspace-stabilized LoRA strategy. Specifically, the projection matrix $A$ is initialized with random Gaussian weights and frozen (shared across all nodes). Each node $k$ optimizes only the projection-back matrix $B_k$. This constraint eliminates the rotation ambiguity inherent in low-rank factorization, ignored by some authors in federated settings~\cite{zhang2024towards,kim2025x}, ensuring that the updates $B_k$ from different nodes share a common basis and can be averaged by the global server as $\theta_{new} = \theta_{fixed} + (\frac{1}{K}\sum B_k) A_{fixed}$. As a second step, we propose weight-decompose geometric adaptation (\textbf{GeoDoRA}) to resolve the magnitude-direction entanglement inherent in standard LoRA~\cite{liu2024dora}. In our unpaired setting this is suboptimal: the direction of the update typically encodes the semantic concept, which must be shared, whereas the magnitude often captures local domain statistics (e.g., image statistics), which should remain private. We will therefore decompose the update into a vector $\mathbf{v}$ and a directional matrix $D$ as

\begin{equation}
    \theta_{new} = \underbrace{\left(\frac{1}{K}\sum v_k \right)}_{v} \odot  \underbrace{ \frac{\theta_{fixed} + (\frac{1}{K}\sum B_k) A_{fixed}}{\| \theta_{fixed} + (\frac{1}{K}\sum B_k) A_{fixed} \|_c}}_{\text{D}},
\end{equation}

where $\odot$ denotes column-wise multiplication, and $\|\cdot\|_c$ is the column-wise norm. \textbf{Our geometric loss $\mathcal{R}_{geo}$ will be applied exclusively to the directional component $D$, enforcing strict semantic alignment across hospitals, while the magnitude $\mathbf{v}$ is allowed to fluctuate freely to absorb domain shifts (e.g. due to different medical scanners and patient cohorts).} 

While standard LoRA and our proposed GeoDoRA utilize a fixed intrinsic rank $r$ across all federation nodes, the informational density of our unpaired modalities varies drastically. For instance, high-dimensional MRI tokens inherently require a higher parameter capacity to capture semantic shifts than sparse tabular clinical data. Enforcing a uniform rank forces unnecessary communication overhead for simpler modalities and risks representational bottlenecks for complex ones. \textbf{To further optimize communication efficiency, we propose dynamically allocating the rank budget for the local directional update matrix $B_{k}$ based on both intrinsic modality complexity and real-time geometric alignment.} To prevent dimensionality mismatches, a node assigned a dynamic rank $r_{k}^{(t)}$ simply truncates the projection matrix $A_{fixed}$ to its first $r_{k}^{(t)}$ rows. At aggregation, the $B_k$ matrix is zero padded to full rank.





\begin{wrapfigure}[16]{r}{0.5\textwidth}
  \begin{center}
    \vspace{-30pt}        \includegraphics[width=0.45\textwidth]{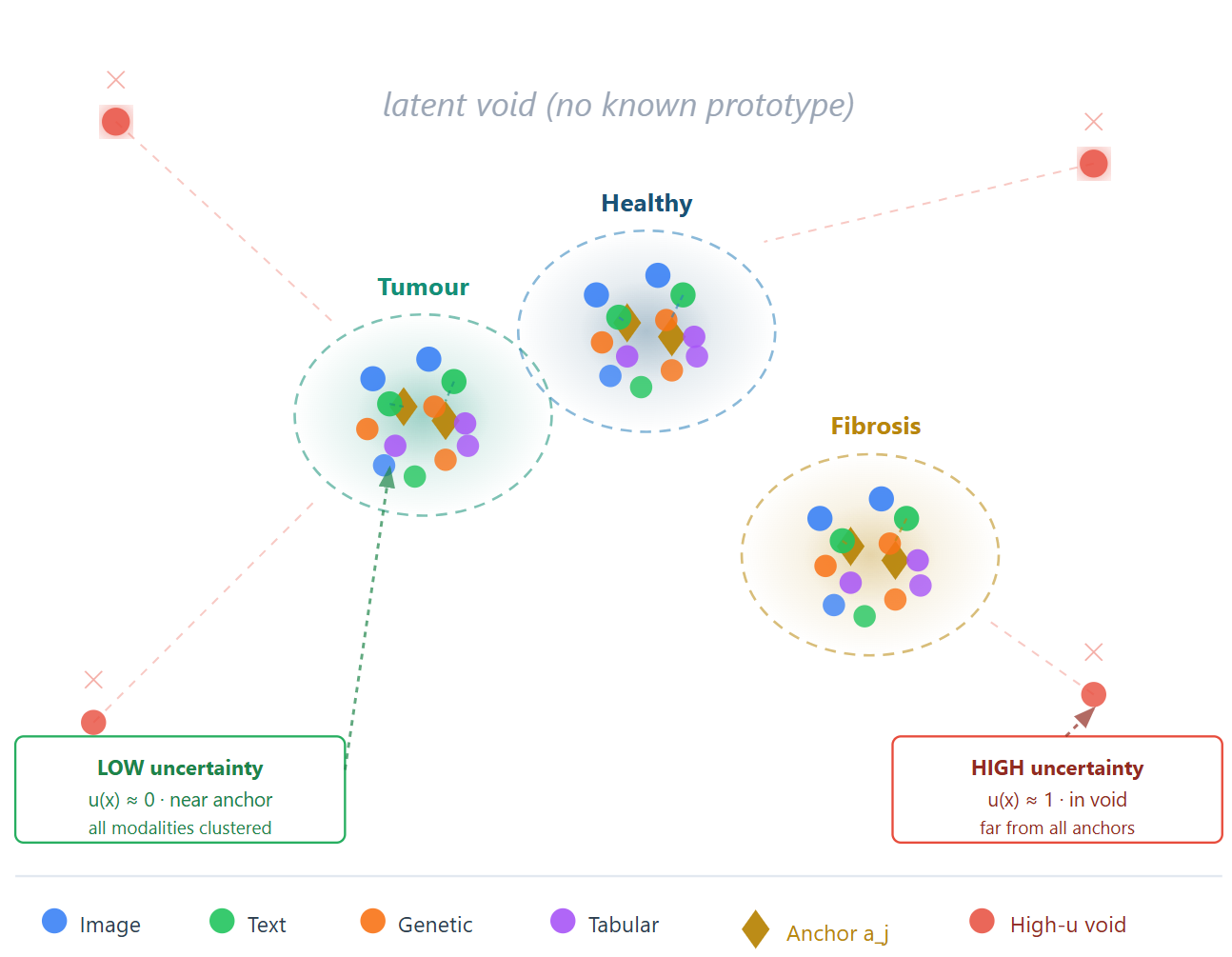}
  \end{center}
  \vspace{-15pt}
  \caption{Efficiently estimating node uncertainty by measuring distances to the anchor dataset in latent space.}
  \label{fig:lap}
\end{wrapfigure}

\textbf{Idea 3: Precision weighted aggregation.} Standard aggregation methods like FedAvg~\cite{mcmahan2017communication} treat all node updates as equally valid. In our unpaired setting, this is flawed: nodes with bad geometric alignment should not influence the global model strongly. We therefore propose a \textbf{precision-weighted aggregation} protocol for the low rank matrices. Standard methods for estimating epistemic uncertainty (e.g. Bayesian ensembles) are computationally prohibitive in a federated setting with limited node resources. To address this, we propose \textit{latent anchor-proximity} (LAP), a deterministic and computationally efficient uncertainty metric that exploits the geometric structure of the proposed CKA alignment, see Figure~\ref{fig:lap}. Since the global consensus kernel $\overline{G}$ represents the shared geometry of trusted medical concepts defined by the anchor set $\mathcal{A}$, we interpret the distance of a local sample to this anchor manifold as a proxy for epistemic uncertainty. We reuse the anchor samples $a_j$, used for the geometric loss, to estimate uncertainty for data sample $x_i$ at node $k$

\begin{equation}
    u\left(x_{i}^{(k)}\right) = \frac{1}{2} \left( 1 - \max_{j \in \{1,\dots,B\}} 
\kappa \left( \text{Pool} \left( f_\theta \left( W_{m_k} \phi_{m_k} \left(x_i^{(k)} \right) \right) \right), \quad \text{Pool} \left(f_\theta \left( W_{m_k} \phi_{m_k}(a_j) \right) \right) \right) \right),    
\end{equation}

where $\kappa(\cdot,\cdot)$ is the cosine similarity. This metric effectively measures the density of the latent space; samples projecting into voids far from any known biological prototype yield $u(x) \approx 1$, indicating high uncertainty. This method incurs negligible computational cost as it utilizes the pre-computed anchor embeddings. Each node weight $p_k$ is then computed as the mean inverse uncertainty $u^{-1}$ over the local dataset according to $p_k = \frac{1}{E} \sum_{i=1}^{N_k}  {u^{-1}\left(x_{i}^{(k)}\right)} $, where $E$ is a normalization factor and $N_k$ is the number of samples for node $k$.  This ensures that the global transformer model is evolved primarily by nodes whose local data is geometrically consistent with the shared biological reality. 

However, applying a single scalar weight $p_{k}$ to the entire local update neglects the dual nature of our GeoDoRA decomposition. Because the magnitude vector $\textbf{v}$ absorbs local domain shifts and the directional matrix $D$ (through $B_k$) encodes shared semantic concepts, we propose a decoupled, component-aware aggregation protocol as a secondary step. The magnitude vector $\textbf{v}$ will be aggregated using the LAP-derived weights $p_{k}$, ensuring the global model safely absorbs diverse domain statistics from confident nodes. Conversely, the semantic directional matrix $D$ (through $B_k$) will be aggregated using a dynamic \textbf{geometric consistency score}, $q_{k}$, computed as

\begin{equation}
q_{k} = \frac{\exp\left(\mathrm{CKA}(G^{(k)}, \overline{G}) / \tau\right)}{\sum_{j=1}^{K} \exp\left(\mathrm{CKA}(G^{(j)}, \overline{G}) / \tau\right)},
\label{eq:geo_consistency}
\end{equation}

where $\tau$ is a temperature hyperparameter controlling the harshness of the penalty. By directly coupling the directional aggregation weight $q_{k}$ to the CKA alignment objective, we create a self-reinforcing feedback loop: nodes that successfully align their local geometry to the global consensus $\overline{G}$ are granted proportionally higher influence over the global semantic direction, while geometrically divergent nodes are naturally quarantined from disrupting the shared transformer manifold. Furthermore, applying a single scalar weight $q_{k}$ to a huge transformer model neglects the hierarchical nature of deep representations. Different modalities inherently require varying network depths to reach semantic consensus; for instance, dense image tokens may require more layers to align with abstract clinical concepts than structured tabular data. To account for this, we propose a depth-aware aggregation protocol that computes layer-specific consensus weights. Rather than evaluating CKA solely at the final layer, we compute intermediate Gram matrices at each transformer block $l \in \{1, \ldots, N_{layers}\}$ and partition the blocks into $S$ consecutive groups $\mathcal{G}_1, \ldots, \mathcal{G}_S$ of size $\lfloor N_{layers}/S \rfloor$. The group-level Gram matrix is obtained as

\begin{equation}
G^{(k,s)} = \frac{1}{|\mathcal{G}_s|} \sum_{l \in \mathcal{G}_s} G^{(k,l)}.
\end{equation}

The global server maintains a group-wise consensus kernel $\bar{G}^{(s)} = \frac{1}{K}\sum_k G^{(k,s)}$, allowing us to define a group-specific aggregation weight $q_k^{(s)}$ for node $k$ and layer group $s$ according to Eq 8. The global model is then aggregated group-by-group. This formulation has two natural special cases: $S = N_{layers}$ recovers full layer-wise aggregation, and $S = 1$ reduces to the standard full model aggregation approach. By treating $S$ as a tuneable hyperparameter, the communication cost per round is an explicit and controllable $K \times S \times B^2$ scalars. This mechanism allows the federation to dynamically trust a modality's structural feature extraction in early layer groups, while gracefully discarding its deeper-layer updates if its semantic manifold diverges from the global biological consensus.

\section*{Preliminary Results} 

\begin{wrapfigure}[15]{l}{0.475\textwidth}
  \begin{center}
    \vspace{-20pt}
    \includegraphics[width=0.475\textwidth]{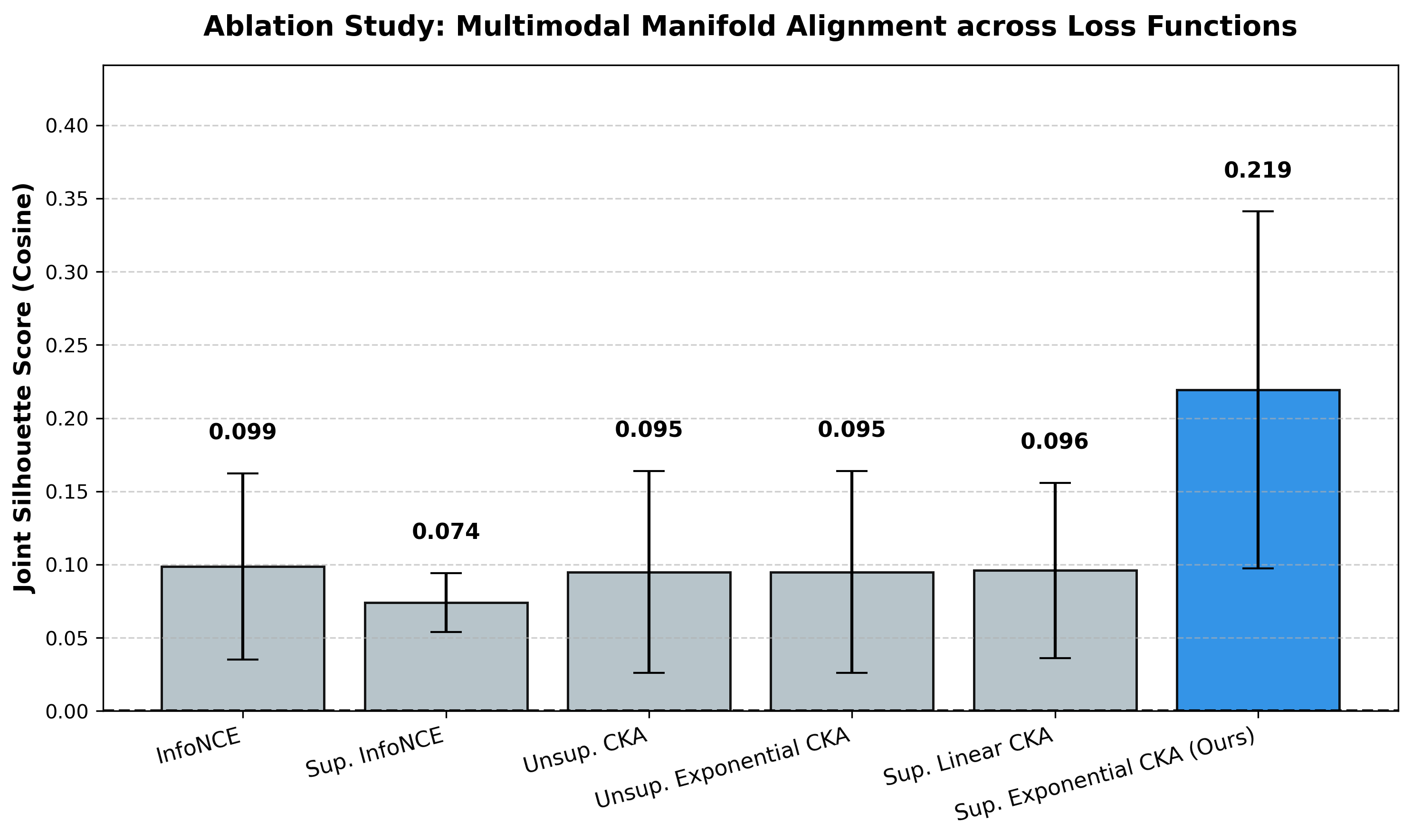}
  \end{center}
  \vspace{-15pt}
  \caption{Comparing different modality alignment losses when aligning image features and tabular features to the same global latent space. The bars represent the mean and standard deviation over 400 hyper parameter combinations. Sup = supervised.}
  \label{results}
\end{wrapfigure}

As a first test of our proposed modality alignment loss, we used paired data from 1073 femur fracture patients (radiographic images and 45 clinical variables) previously used to train \textbf{multimodal} classifiers~\cite{schilcher2024fusion}. The dataset is imbalanced with 85\% normal fractures and 15\% atypical fractures. We used a pre-trained and frozen ResNet18 CNN to obtain image feature vectors (dimension 512), and used 200 of the patients to train a multi-task autoencoder for tabular data to generate a feature vector of dimension 16. We used data from 400 patients for training (anchor dataset), and 200 patients as validation data (early stopping). To demonstrate that our proposed approach learns generalizable disease representations rather than statistical shortcuts, we evaluated the global latent space using the joint silhouette score on the remaining 273 patients. The silhouette score measures the macroscopic geometric separation of classes within the hyper sphere. We compared our proposed CKA loss and several modifications for training adapters to project the multimodal features into a joint global latent space against the common InfoNCE loss~\cite{oord2018representation}. A grid search was used over learning rate, global dimension, weight decay, and temperature scaling. Averaged over 400 hyper parameter combinations, the proposed supervised exponential CKA loss outperformed the other approaches, see Figure~\ref{results}. 

\clearpage
\newpage
\pagestyle{fancy}
\fancyhead[L]{References}

\printbibliography

\end{document}